\documentclass[10pt,twocolumn,letterpaper]{article}

\usepackage{iccv}
\usepackage{times}
\usepackage{epsfig}
\usepackage{graphicx}
\usepackage{amsmath}
\usepackage{amssymb}
\usepackage{bbding}
\usepackage{multirow}
\usepackage{cite}
\usepackage{colortbl}
\usepackage{booktabs}
\usepackage{makecell}
\usepackage[table]{xcolor}

\usepackage[pagebackref=true,breaklinks=true,letterpaper=true,colorlinks,bookmarks=false]{hyperref}

\iccvfinalcopy 

\def\iccvPaperID{9815} 

\ificcvfinal\fi

\begin{document}

\title{Point Contrastive Prediction with Semantic Clustering for \\Self-Supervised Learning on Point Cloud Videos}

\author{
Xiaoxiao Sheng$^{1}$\footnotemark[1]\ , \ \ 
Zhiqiang Shen$^{1}$\footnotemark[1]\ , \ \ 
Gang Xiao$^{1}$\footnotemark[2], \ \ 
Longguang Wang$^{2}$, \ \ 
Yulan Guo$^{3}$, \ \  
Hehe Fan$^{4}$ \ \ \\
$^1$Shanghai Jiao Tong University\ \ \ \ \ 
$^2$Aviation University of Air Force\\
$^3$Sun Yat-sen University\ \ \ \ \ \ \ \ \ \ \ 
$^4$Zhejiang University\ \ \ \ \ \\
{\tt\small \{shengxiaoxiao, shenzhiqiang, xiaogang\}@sjtu.edu.cn}
}

\maketitle
\ificcvfinal\thispagestyle{empty}\fi

\renewcommand{\thefootnote}{\fnsymbol{footnote}}
\footnotetext[1]{These authors contributed equally.}
\footnotetext[2]{Corresponding author.}

\begin{abstract}
We propose a unified point cloud video self-supervised learning framework for object-centric and scene-centric data. Previous methods commonly conduct representation learning at the clip or frame level and cannot well capture fine-grained semantics. 
Instead of contrasting the representations of clips or frames, in this paper, we propose a unified self-supervised framework by conducting contrastive learning at the point level. Moreover, we introduce a new pretext task by achieving semantic alignment of superpoints, which further facilitates the representations to capture semantic cues at multiple scales. 
In addition, due to the high redundancy in the temporal dimension of dynamic point clouds, directly conducting contrastive learning at the point level usually leads to massive undesired negatives and insufficient modeling of positive representations. To remedy this, we propose a selection strategy to retain proper negatives and make use of high-similarity samples from other instances as positive supplements.
Extensive experiments show that our method outperforms supervised counterparts on a wide range of downstream tasks and demonstrates the superior transferability of the learned representations. 

\end{abstract}

\section{Introduction}

Point cloud videos captured by 3D sensors describe the dynamics of objects and their surrounding environments, and have been applied in a wide range of fields to perceive the environment, including robotics and autonomous driving.
Early point cloud understanding approaches mainly focus on the geometric modeling of static point clouds \cite{Chen_2022_CVPR, hu2021learning, zhang2022not}.
Recently, more attention has been paid to point cloud videos \cite{fan2022point, PST2, wen2022point, fan2021deep}. 
However, since obtaining point-wise annotation for  point cloud videos is labor-intensive \cite{xie2020pointcontrast, afham2022crosspoint}, conducting self-supervised learning on dynamic point clouds has drawn increasing interest.
Despite the great success of recent self-supervised learning on images and static point clouds \cite{chen2020simple, he2020momentum, grill2020bootstrap, zbontar2021barlow, xie2022simmim}, 
two questions still remain for point cloud videos:

\begin{figure}[t]
	\centering
	\includegraphics[width=1\columnwidth]{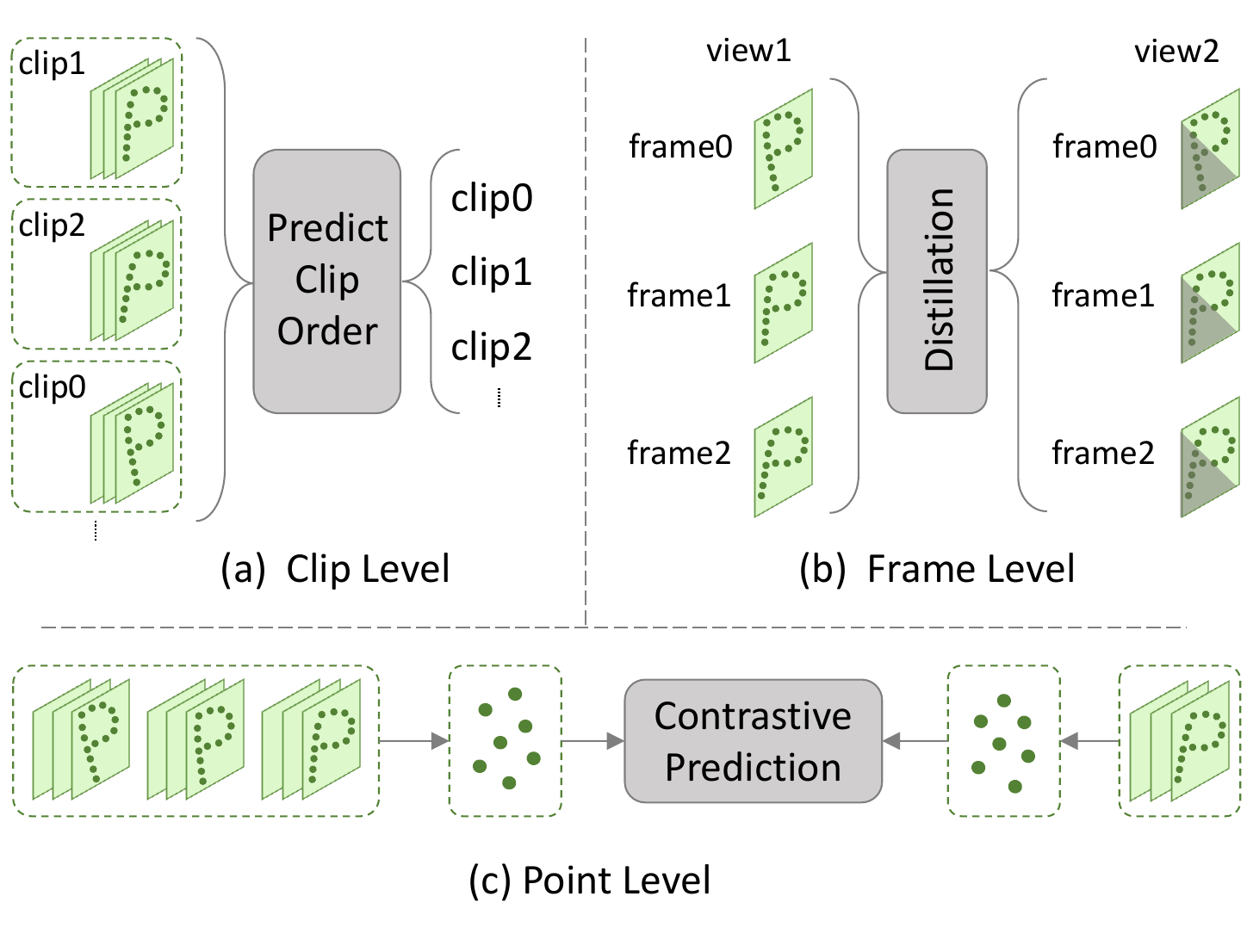}
	\caption{Existing works utilize clip-level (a) or frame-level (b) instances for point cloud video pre-training, while we focus on point-level (c) pre-training.}
	\label{fig1}
\end{figure}


(\romannumeral1) \emph{How to build a unified self-supervised framework?}
Multi-granularity perception of point cloud videos is demanded in different tasks, such as classification, semantic segmentation, and part segmentation. Existing works conduct self-supervised learning by predicting the orders of randomly shuffled clips or distilling spatiotemporal knowledge based on complete-to-partial sequences \cite{wang2021selfwacv, dong2022complete} (Fig.~\ref{fig1}(a-b)). 
Representations learned by these paradigms focus more on frame-level semantics and cannot well capture fine-grained semantic cues.
Therefore, building a unified self-supervised framework that can learn representation rich in multi-granularity semantics is highly demanded.

(\romannumeral2) \emph{How to achieve effective learning between local samples?} 
To build a unified self-supervised framework for multiple point cloud video tasks, it is necessary to learn fine-grained semantics at the level of local samples.
Traditional contrastive learning constructs two views from the same instance as positives and pushes away all other instances \cite{chen2020simple, he2020momentum, caron2020unsupervised, caron2021emerging, grill2020bootstrap, zbontar2021barlow, wang2022importance, chen2021exploring}. 
Since dynamic point clouds are highly redundant in the temporal dimension, directly applying previous approaches to local samples may introduce massive undesired negatives.
Therefore, how to conduct effective learning on local samples to obtain fine-grained semantics still remains under-investigated.

In this paper, we propose a unified point-based contrastive prediction framework, termed as PointCPSC, for self-supervised learning on point cloud videos.
We conduct representation learning at the point level by contrasting local superpoints of predictions and targets (Fig.~\ref{fig1}(c)).
Regarding challenge (\romannumeral1), we propose a new pretext task to 
{align the predicted prototypes and target prototypes, as well as  soft category assignments between predictions and targets.}
For challenge (\romannumeral2), we propose a negative sample selection strategy and employ higher similar samples from other instances as positive supplements.
Compared with the frame-based self-supervised framework, our method achieves more effective representation modeling at a finer granularity, and can be applied to multiple point cloud video understanding tasks.
The main contributions of our paper are summarized as follows:
\begin{itemize}
\item We propose a unified self-supervised contrastive learning framework for point cloud videos. Our framework facilitates the representations to capture both fine-grained dynamics and hierarchical semantics for multiple downstream tasks.

\item We introduce a new pretext task by achieving the semantic alignment {between predictions and targets.}
This facilitates our self-supervised framework to capture semantic information on multiple scales.

\item {We design a feature similarity based sample selection strategy to retain proper negatives and positive neighbors for effective representation learning. }

\item Our framework produces remarkable performance on a wide range of downstream tasks. We also perform extensive ablation studies and visualized analysis to demonstrate the effectiveness of our method.

\end{itemize}

\section{Related Work}

In this section, we first present related works of contrastive learning on images and static point clouds.
Then, we introduce the advanced works about dynamic point cloud modeling.
\subsection{Contrastive Learning}
Self-supervised learning has achieved great success in images, notably represented by instance-based discriminative methods \cite{chen2020simple, chen2020big, he2020momentum, caron2020unsupervised, caron2021emerging, grill2020bootstrap, chen2021exploring, zbontar2021barlow, Wang_2021_CVPR}.
This classic paradigm augments two views of an instance as a positive pair, while treating all views of other instances as negatives.
Many techniques have been introduced to enhance the representation learning capability \cite{wang2020understanding, tao2022exploring, wang2022importance, chen2021exploring, grill2020bootstrap, caron2021emerging}.
He \etal~\cite{he2020momentum} introduced dynamic queues to store massive negatives. 
Chen \etal~\cite{chen2021exploring} indicated that massive negatives and momentum updated encoder are not essential for contrastive learning, and a simple siamese network structure with a stop gradient can avoid mode collapse.
Caron \etal~\cite{caron2021emerging} established a teacher-student self-distillation framework and aligned the two branches with a classification loss.
In addition, Debidatta \etal \cite{dwibedi2021little} utilized feature similarity to mine the nearest neighbors from the support set as positive sample supplements, making positive representations robust and invariant to deformations.

Recently, contrastive learning has been extended to static point cloud understanding.
PointContrast \cite{xie2020pointcontrast} generates two views of point clouds, and then utilizes the contrastive loss to pull matched point pairs and push unmatched ones.
DepthContrast \cite{zhang2021self} learns global representations from two augmented depth views by setting an instance discrimination task.
Although contrastive learning has achieved great success on images and static point clouds, the utilization of contrastive learning on dynamic point clouds is still under-investigated.

\begin{figure*}[ht]
        \centering
        \includegraphics[width=1\linewidth]{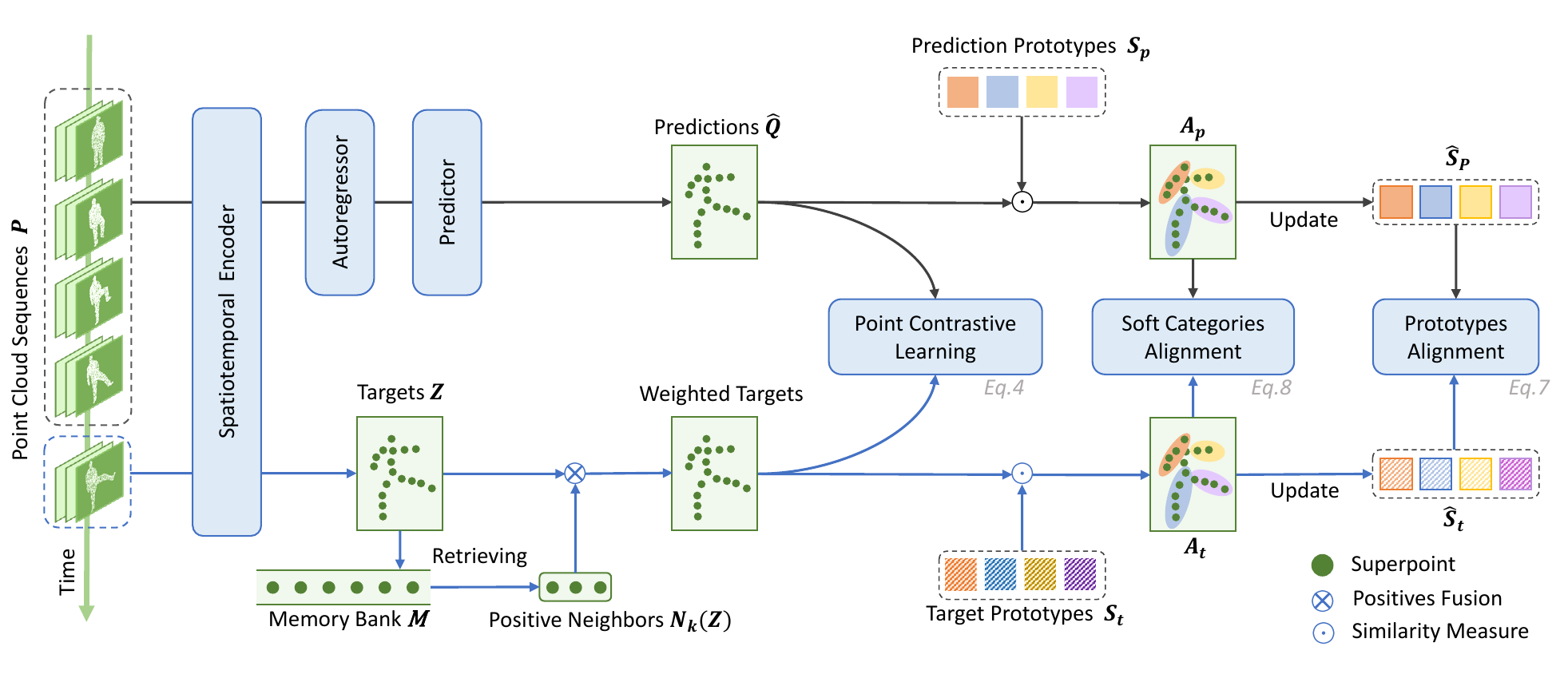}
        \caption{The framework of our PointCPSC. {The  samples with lower similarities in the memory bank are used as negatives. We only present the selection of positive neighbors from the memory bank for simplicity.}
        }
        \label{fig2}
\end{figure*}

\subsection{Dynamic Point Cloud Modeling
}

Currently, most point cloud video understanding methods focus on supervised learning \cite{p4d, fan2022point, PST2, wen2022point, MeteorNet, pstnet, fan2021deep, 3dv, zhong2022no}.
Liu~\etal \cite{MeteorNet} added a temporal dimension to PointNet++ \cite{pointarxiv} to process dynamic point clouds.
Wang~\etal \cite{3dv} extracted motion information from regularized voxels, and then combined these voxels with raw points for spatiotemporal modeling.
Fan~\etal \cite{pstnet} used stacked convolutions to extract hierarchical spatiotemporal features.
P4Transformer \cite{p4d} captures long relationships between tokens obtained from spatiotemporal tubes.
Zhong \etal \cite{zhong2022no} and Wen \etal \cite{wen2022point} introduced traditional techniques, such as ST-surface or primitives, into the existing network structure to effectively learn spatiotemporal representations.
Niemeyer \etal \cite{niemeyer2019occupancy} learned a temporally and spatially continuous vector field to assign a motion vector to each point, which is suitable for generative tasks such as dynamic point cloud reconstruction.
Rempe \etal \cite{rempe2020caspr} learned object-centric spatiotemporal representations from normalized point clouds and proved to be effective on multiple downstream tasks.

Meanwhile, several works make attempts to conduct self-supervised learning on dynamic point clouds.
Wang~\etal\cite{wang2021selfwacv} divided input sequences into several temporal clips and then predicted the correct order of randomly shuffled clips.
Dong~\etal\cite{dong2022complete} used complete and partial sequences as inputs to the teacher and student networks for realizing spatiotemporal knowledge distillation.
However, the representations learned by contrasting samples at the clip or frame level cannot capture local spatiotemporal dynamics.
{To remedy this, Sheng~\etal\cite{pointcmp} proposed a self-supervised framework to learn fine-grained representations through contrastive prediction and reconstruction.
Although spatiotemporal reconstruction of raw points pays more attention to fine-grained information, the learned representations are susceptible to noises and  the network is difficult to be optimized.
Shen~\etal\cite{pointcpr} combined contrastive learning and masked predictions to achieve self-supervised representation learning.
However, their clip-level masking strategy is insufficient to explore fine-grained dynamics of point clouds. 
Different from the above methods, in this paper, we propose to conduct contrastive learning between superpoints and introduce semantic clustering as a pretext task to learn representations versatile to diverse downstream tasks.
}



\section{Method}

The overall framework of our PointCPSC is presented in Fig.~\ref{fig2}.
A point cloud video is denoted as $\boldsymbol{P}\in\mathbb{R}^{T\times N\times3}$, where $T$ is sequence length and $N$ is the number of points in each frame.
We equally divide the video into $L$ segments.
After all segments are processed by the spatiotemporal encoder, the former $L$-1 segments are fed into transformer autoregressor to predict the $L$-th target segment in the latent space.
We follow previous works \cite{he2020momentum, chen2021exploring} to implement a predictor to further transform the predictions.
To make local contrasts more effective and model comprehensive positive representations, we propose to select appropriate negatives and beneficial positive neighbors.
Meanwhile, we also perform semantic clustering to adapt the self-supervised framework for multiple downstream tasks.

\subsection{Point Contrastive Prediction}

Following \cite{pstnet}, spatiotemporal tubes are defined as tubes within spatial radius $s$ and temporal radius $t$ centered on certain points.
We first encode these spatiotemporal tubes to obtain the embeddings of superpoints.
{These superpoints aggregate local information and can well preserve local semantics, thereby facilitating the learning of fine-grained information}.

Specifically, after the $L$-th segment is encoded by the spatiotemporal encoder, we obtain the target embeddings $\boldsymbol{Z}\in\mathbb{R}^{l \times r \times c}$, where $l$, $r$, and $c$ are frame length, superpoint number, and feature dimension, respectively.
We then take the representations of the $L$-1 segment as predictions, which are denoted as $\boldsymbol{Q}\in\mathbb{R}^{l \times r \times c}$.
However, owing to the disorder of point clouds, the predictions are not aligned with the corresponding targets.
We take target positions as anchors to search for neighbors within predictions and perform feature interpolation to obtain updated predictions $\boldsymbol{\hat Q}\in\mathbb{R}^{l \times r \times c}$.

\textbf{Negative Selection.}
Due to the high redundancy in a point cloud video, numerous superpoints contain similar semantics.
For effective contrastive learning, samples with high similarities are discarded.
Specifically, we use dynamically updated memory bank, denoted as $\boldsymbol{M}$, to store history target embeddings.
During per-training, we calculate the similarities between the current target and those in $\boldsymbol{M}$ as follows:
\begin{equation}
   sim = \cos(\boldsymbol{m},\boldsymbol{z}), \boldsymbol{m}\in \boldsymbol{M}, \ \ 
\end{equation}
\noindent where $sim$ represents the similarity between history embedding $\boldsymbol{m}$ and current superpoint $\boldsymbol{z}$. 
For embeddings in $\boldsymbol{M}$, we sort their similarities with $\boldsymbol{z}$ and retain 70\% negatives with {the lowest} similarity for contrastive learning.

\textbf{Positive Neighbors.}
Instead of directly employing the embeddings of the same spatiotemporal position as positives,
we propose to explore favorable positive neighbors by utilizing feature similarity:
\begin{equation}
   N_k(\boldsymbol{z}) = \mathop{\mathrm{argmax}}\limits_{\boldsymbol{m}\in\boldsymbol{M}}{(\cos(\boldsymbol{m},\boldsymbol{z}), {\rm top}_n = K)},
\end{equation}
\noindent where $N_k(\boldsymbol{z})$ represents the retrieved $K$ neighbors related to the current target superpoint $\boldsymbol{z}$. 
Following \cite{neighborICLR}, we adaptively introduce these positive neighbors into local Info Noise Contrastive Estimation (InfoNCE) loss \cite{oord2018representation} to perform contrastive learning between predictions and targets, which is represented as follows:
\begin{equation}
    \boldsymbol{w} = \mathrm{Softmax}(N_k( \boldsymbol{z})\cdot \boldsymbol{z})\in \mathbb{R}^{K},
\label{eq.3}
\end{equation}
\begin{equation}
\centering
\small
\mathcal{L}_{l}=-\log\frac{\sum_{j=0}^{K}w_j\exp(\boldsymbol{z}^T \boldsymbol{q}/ \tau)}{\sum_{j=0}^{K}w_j\exp(\boldsymbol{z}^T \boldsymbol{q}/\tau)\!+\!\sum_{\boldsymbol{q}^{'}\in{\Psi}}{\exp{(\boldsymbol{z}_i^T \boldsymbol{q}^{'}/\tau)}}},
\end{equation}
where $\boldsymbol{q}\in\{\boldsymbol{\hat q}_+ \cup N_k(\boldsymbol{z})\}$ is the positive set that contains the positive sample $\boldsymbol{\hat q}_+$ and neighbors $N_k(\boldsymbol{z})$, $\Psi$ is the negative set, {$w_0$ is set as 1 and means the weight between $\boldsymbol{z}$ and $\boldsymbol{\hat q}_+$, $w_{j=1,\ldots,K}$ is calculated using Eq.\ref{eq.3}}, and $\tau$ is temperature hyper-parameter.

Overall, we make point contrastive prediction more effective by selecting proper negatives, and robust representations are learned by supplying positive neighbors from other instances.
The feature similarity is utilized as adaptive weights to combine neighbors and positive samples.
We further investigate how to utilize retrieved positive neighbors in ablation studies.

\subsection{Semantic Clustering}

Local spatiotemporal representations are learned based on the above contrastive learning framework.
As multiple-granularity semantics are critical to diverse downstream tasks, we introduce a semantic clustering task on dynamic point clouds.

Specifically, we first parameterize two group prototypes for prediction and target embeddings.
During pre-training, these two group prototypes are gradually learned.
The distances from predictions to their corresponding prototypes are calculated to obtain soft category distributions for each predicted superpoint.
The same operation is also performed to obtain soft category distributions for each target superpoint.
Intuitively, the embeddings of predictions and targets should follow the same category probability distribution.
Meanwhile, the two group prototypes should also follow approximate distributions. 
We denote the initial target prototypes as $\boldsymbol{S}_t = [\boldsymbol{s}_1, \boldsymbol{s}_2,\ldots, \boldsymbol{s}_k]\in\mathbb{R}^{k \times c} $ and the semantic clustering is achieved as follows:
\begin{equation}
\boldsymbol{A}_t = \underset{k}{\rm{Softmax}}(\boldsymbol{Z}\cdot \boldsymbol{S}_t^T) \in \mathbb{R}^{l \times r \times k},
\end{equation}
\begin{equation}
\boldsymbol{\hat S}_t = \frac{1}{\sum_{i,j}\boldsymbol{A}_{t}[i,j]}{\sum_{i,j}}\boldsymbol{A}_t[i,j]\odot\boldsymbol{Z}[i,j]\in\mathbb{R}^{k \times c},
\end{equation}
\noindent where $\boldsymbol{A}_t$ represents soft category assignments of $\boldsymbol{Z}$, $\boldsymbol{\hat S}_t$ are updated prototypes of $\boldsymbol{Z}$, and $\odot$ is a Hadamard product.
Similarly, the soft category distributions $\boldsymbol{A}_p$ and updated prototypes of predictions $\boldsymbol{\hat S}_p$ can be obtained.

Following \cite{wen2022self}, an extra predictor is utilized to further transform $\boldsymbol{\hat S}_p$.
Finally, an InfoNCE loss \cite{oord2018representation} is employed to align $\boldsymbol{\hat S}_p$ and $\boldsymbol{\hat S}_t$ as follows:
\begin{small}
\begin{equation}
\mathcal{L}_{c} = -\log\!\frac{\exp(\boldsymbol{\hat s}_t^T \boldsymbol{\hat s}_p/ \tau)}{\exp(\boldsymbol{\hat s}_t^T \boldsymbol{\hat s}_p/\tau)\!+\!\sum_{\boldsymbol{\hat s}_p^{'}\in{\phi}}{\exp{(\boldsymbol{\hat s}_t^T \boldsymbol{\hat s}_p^{'}/\tau)}}}, 
\end{equation}
\end{small}
 
\noindent where $(\boldsymbol{\hat s}_t, \boldsymbol{\hat s}_p)$ is a positive pair, and $\phi$ is the negative set that contains unmatched prototypes.

Moreover, we utilize Kullback-Leibler Divergence (KL) loss to achieve the alignment of soft category distributions between predictions and targets:
\begin{equation}
\mathcal{L}_k = \sum_{i=1}^{k}{ \boldsymbol{ a}}_p^i(\log\boldsymbol{a}_p^i-\log\boldsymbol{a}_t^i),
\end{equation}
where $\boldsymbol{a}_p^i$ and $\boldsymbol{ a}_t^i$ are $i$-th category probabilities of predictions and targets, respectively.

Overall, the total loss of our self-supervised framework consists of three parts:
\begin{equation}
    \mathcal{L}_{total} = \mathcal{L}_{l} + \lambda_1\mathcal{L}_{c} + \lambda_2 \mathcal{L}_{k}, 
\end{equation}
where $\lambda_1$ and $\lambda_2$ are hyper-parameters for balance.
By performing the alignment of superpoint categories and prototypes, our method can well capture multiple-granularity semantics.

Note that, prototypes and soft category alignments are performed when pre-training on dynamic point cloud semantic segmentation. This is because soft category alignment is capable of extracting fine-grained point-level information which is beneficial for segmenting objects.
While pre-training on action recognition, we only conduct prototypes alignment to provide high-level semantics.

\section{Experiments}

Firstly, the dataset benchmarks and implementation details are introduced, and then we compare the performance of PointCPSC with previous methods under multiple settings.
Extensive ablation studies are also conducted to demonstrate the effectiveness of each sub-module in our framework.
Finally, we present qualitative analysis and visualizations to verify our motivation.

\subsection{Datasets and Pre-training Details}

We perform point cloud action recognition on MSRAction-3D \cite{msr} and NTU-RGBD \cite{ntu60}, 4D semantic segmentation on Synthia 4D \cite{minke}, and gesture recognition on NvGesture \cite{molchanov2016online}.

The MSRAction-3D \cite{msr} dataset records 567 human action sequences with Kinect, including 20 action categories performed by 10 subjects.
We follow \cite{MeteorNet} to obtain 270 training videos and 297 test videos.

The NTU-RGBD \cite{ntu60} dataset collects 56880 videos recorded by three cameras from different angles, with a total of 40 subjects and 60 categories.
There are 40,320 training videos and 16,560 test videos under a cross-subject setting.

The Synthia 4D dataset \cite{minke} contains 6 videos of different driving scenarios, generated from the Synthia dataset \cite{ros2016synthia}.
Following \cite{minke, MeteorNet}, this dataset is split into 19,888 training frames, 815 validation frames, and 1,886 test frames.

The NvGesture \cite{molchanov2016online} dataset collects 1532 dynamic sequences with 25 categories.
We follow \cite{min2020efficient} to obtain 1050 training videos and 482 test videos.

\begin{table}[t]
    \centering
    \footnotesize
    \caption{Action recognition accuracy (\%) on MSRAction-3D.}
    \setlength{\tabcolsep}{1.5mm}
    \begin{tabular}{l|l|cccc}
    \toprule
    \multicolumn{2}{c|}{\multirow{2}{*}{\textbf{Methods}}} & \multicolumn{4}{c}{\textbf{\#Frames}}\\
    \cmidrule(r){3-6}
    \multicolumn{2}{l|}{}     & 8     & 12    & 16     & 24 \\
    \midrule
    \multirow{8}{*}{{\makecell[l]{Supervised\\Learning}}} &
    MeteorNet~\cite{MeteorNet}  & 81.14 & 86.53 & 88.21 & 88.50 \\
    & Kinet~\cite{zhong2022no}  & 83.84 & 88.53 & 91.92 & 93.27 \\
    & PST$^2$~\cite{PST2}       & 86.53 & 88.55 & 89.22 & -     \\
    & PPTr~\cite{wen2022point}  & 84.02 & 89.89 & 90.31 & 92.33 \\
    & P4Transformer~\cite{p4d}   & 83.17 & 87.54 & 89.56 & 90.94 \\
    & PST-Transformer~\cite{fan2022point}   & 83.97 & 88.15 & 91.98 & \textbf{93.73} \\
    & \cellcolor{gray!20}PSTNet~\cite{pstnet} & \cellcolor{gray!20}83.50 & \cellcolor{gray!20}87.88 & \cellcolor{gray!20}89.90 & \cellcolor{gray!20}91.20 \\
    & PSTNet++~\cite{fan2021deep}  & 83.50 & 88.15 & 90.24 & 92.68 \\    
    \midrule
    {\makecell[l]{End-to-end \\Fine-tuning}} \ \ \ & \textbf{PointCPSC}  & \textbf{88.89} & \textbf{90.24} & \textbf{92.26} & 92.68 \\
    \midrule
    {\makecell[l]{Linear \\ Probing}}  & \textbf{PointCPSC} \ \ \  & 86.87 & 89.56 & 88.89 & 90.24 \\
    \bottomrule
    \end{tabular}
    \label{MSRAction-3D}
\end{table}

\textbf{Pre-training on action recognition.} PSTNet \cite{pstnet} is utilized as our encoder to conduct experiments.
{During training, we use 88 clips as one batch, where each clip contains 24 1024-point frames.}
The frame interval of sampling for MSRAction-3D and NTU-RGBD are set to 1 and 2, respectively.
The number of neighbors for the ball query is set to 9.
The spatial search radius is set to 0.5 and 0.1 for MSRAction-3D and NTU-RGBD, respectively.
Following \cite{pstnet}, random scaling is adopted for data augmentation. The AdamW optimizer \cite{adamw} with a cosine decay scheduler is employed for optimization.
We pre-train the model for 200 epochs with an initial learning rate of 0.0008.
The temperature hyper-parameter is set to 0.01.

\textbf{Pre-training on semantic segmentation.} 
The encoder in P4Transformer \cite{p4d} is adopted to conduct experiments.
The 04 sequence of the Synthia 4D dataset is employed for pre-training.
We sample 4-frame clips with each frame containing 4096 points for training.
The frame interval is set to 1, and the spatial search radius and the number of neighbors for the ball query are set to 0.9 and 32, respectively. The data augmentation strategy in \cite{p4d} is adopted in the experiments.
We employ the same optimization strategies as those for pre-training on action recognition.

\begin{table*}[t]
    \centering
    \small
    \setlength{\tabcolsep}{0.85mm}
    \caption{Semantic segmentation accuracy (\%) on the Synthia 4D dataset.}
    \begin{tabular}{l|c|c|cccccccccccc|c}
    \toprule
    \text{Methods} & \text{Input} & \text{Frame} &\text{Bldn} &\text{Road}  &\text{Sdwlk} &\text{Fence} &\text{Vegittn} &\text{Pole} &\text{Car} &\text{T. Sign} &\text{Pedstrn} &\text{Bicycl} &\text{Lane} &\text{T. Light} & \text{mIOU}\\
    \midrule
    Minkowski\cite{minke} & voxel & 3 &  90.13 & 98.26 & 73.47 & 87.19 & \textbf{99.10} & 97.50 & 94.01 & 79.04 & \textbf{92.62} & 0.00 &50.01 & 68.14 & 77.46 \\
    PointNet++\cite{pointarxiv} & point & 1 & 96.88 & 97.72 & 86.20 & 92.75 & 97.12 & 97.09 &90.85 &66.87 &78.64 &0.00 &72.93 &75.17 &79.35\\
    MeteorNet\cite{MeteorNet} & point & 3 & \textbf{98.10} &97.72 &88.65 &94.00 &97.98 &97.65 &93.83 &84.07 &80.90 &0.00 &71.14 &77.60 &81.80 \\
    PSTNet\cite{pstnet} &point &1 & 96.32 &98.07 &85.40 &94.66 &97.16 &97.51 &94.83 &76.65 &76.99 &0.00 &75.39 &76.45 &80.79\\
    PSTNet\cite{pstnet} &point &3 & 96.91 &98.33 &90.83 &95.00 &96.96 &97.61 &95.15 &77.45 &85.68 &0.00 &75.71 &77.28 & 82.24 \\
    \rowcolor{gray!20}P4Transformer\cite{p4d} &point &1 & 96.76 &98.23 &92.11 &95.23 &98.62 &97.77 &95.46 &80.75 &85.48 &0.00 &74.28 &74.22 &82.41 \\
    \rowcolor{gray!20}P4Transformer\cite{p4d} &point &3 & 96.73 &\textbf{98.35} &94.03 &95.23 &98.28 &98.01 &95.60 &81.54 &85.18 &0.00 &75.95 &79.07 & 83.16 \\
    \midrule
    \textbf{PointCPSC} & point & 3 & 95.88 &98.31 &\textbf{94.13} &\textbf{96.32} &97.12 &\textbf{98.55} &\textbf{95.74} &\textbf{85.35} &87.11 &0.00 &\textbf{78.85} &\textbf{86.28} & \textbf{84.47} \\
    \bottomrule
    \end{tabular}
    \label{Synthia4D}
\end{table*}

\subsection{End-to-end Fine-tuning}
\label{Sec4.2}

We first perform pre-training on MSRAction-3D, and then add a new classifier after the encoder for fine-tuning.
Two linear layers with a batch normalization layer are adopted as the classifier.
Following the previous works \cite{pstnet, p4d, MeteorNet}, we test the performance with various lengths of frames.
2048 points are sampled for each frame.
The spatial search radius and the number of neighbors for the ball query are set to 0.3 and 9, respectively.
We finetune the pre-trained model for 35 epochs and employ a warmup strategy. 
We compare the performance of our PointCPSC with previous supervised methods in Table~\ref{MSRAction-3D}.
As we can see, PointCPSC consistently outperforms the baseline method PSTNet under different frames.
This demonstrates the effectiveness of our method, which helps the model to learn semantic information that is beneficial to the point cloud action recognition task.

After pre-training on Synthia 4D, we follow \cite{p4d,pang2022masked} to add a decoder and a classifier for fine-tuning.
During fine-tuning, 3-frame clips with each frame containing 16384 points are sampled.
The spatial search radius and the number of neighbors for the ball query are set to 0.9 and 32, respectively.
We finetune the pre-trained model for 150 epochs and adopt the warmup strategy.
We compare our PointCPSC with other supervised methods and the results are presented in Table~\ref{Synthia4D}. 
The PointCPSC with 3 frames achieves 84.47 mIOU, which is 2\% higher than that of P4Transformer with 1 frame.
This indicates that temporal context information benefits semantic segmentation. 
Compared with the baseline P4Transformer, PointCPSC achieves significant improvements, especially in small object segmentation, including traffic signs, pedestrians, lanes, and traffic lights.
This validates that our self-supervised framework can well fit fine-grained downstream tasks.

\begin{table}[t]
    \centering
    \small
    \caption{Action recognition accuracy (\%) on NTU-RGBD under cross-subject setting.}
    \setlength{\tabcolsep}{4mm}
    \begin{tabular}{lc}
    \toprule
    \textbf{Methods}    & \textbf{Accuracy (\%)} \\
    \midrule
    3DV-Motion~\cite{3dv} (voxel)            & 84.5 \\
    3DV-PointNet++~\cite{3dv} (voxel+point)  & 88.8 \\
    Kinet~\cite{zhong2022no}                 & 92.3 \\
    P4Transformer~\cite{p4d}                 & 90.2 \\
    PST-Transformer~\cite{fan2022point}      & 91.0 \\
    \rowcolor{gray!20}
    PSTNet~\cite{pstnet}                     & 90.5 \\
    PSTNet++~\cite{fan2021deep}              & 91.4 \\
    \midrule
    \textbf{PointCPSC} (50\% Semi-supervised)&  88.0 \\
    \bottomrule
    \end{tabular}
    \label{NTU}
\end{table}

\subsection{Linear Probing}

After pre-training on MSRAction-3D, we evaluate the pre-trained encoder under the setting of linear probing.
The same experimental setups as fine-tuning are adopted.
As shown in Table~\ref{MSRAction-3D}, our results are competitive compared to previous methods, where the performance of PointCPCS with 8 frames outperforms all supervised methods.
Furthermore, our method with 12 frames achieves the accuracy of 89.56\%,  surpassing the baseline PSTNet \cite{pstnet} with notable margins.
These results demonstrate that our pre-training can learn beneficial high-level semantics.

\subsection{Semi-supervised Learning}

We first pre-train the models on NTU-RGBD, and then conduct semi-supervised fine-tuning with
50\% training data under the cross-subject setting.
The spatial radius is set as 0.5.
We finetune 20 epochs and adopt a warmup strategy.
The other experimental setups and optimization strategies are the same as those used for fine-tuning on MSRAction-3D.
As shown in Table~\ref{NTU}, we compare the performance of PointCPSC with previous supervised methods.
Our method with only 50\% annotated data achieves the accuracy of 88.0\%.
This clearly demonstrates the effectiveness of our self-supervised pre-training, which learns advantageous information to assist in semi-supervision.

\subsection{Transfer Learning}

We conduct pre-training on NTU-RGBD and then transfer the pre-trained encoder to gesture recognition to demonstrate the generalization of the pre-trained representations.
We finetune the pre-trained model on the NvGesture dataset for 50 epochs. During fine-tuning, 32 1024-point frames are sampled.
The batch size and initial learning rate are set to 32 and 0.02, respectively.
The SGD optimizer with cosine decay strategy is adopted for optimization.
We compare our PointCPSC with other supervised methods and the results are shown in Table~\ref{NvGesture}.
It can be seen that our method facilitates the baseline PSTNet to achieve higher accuracy.
This validates that our method has superior generalization capability and the learned representations are beneficial for gesture recognition on point cloud videos.

\begin{figure*}[ht]
        \centering
        \includegraphics[width=1\linewidth]{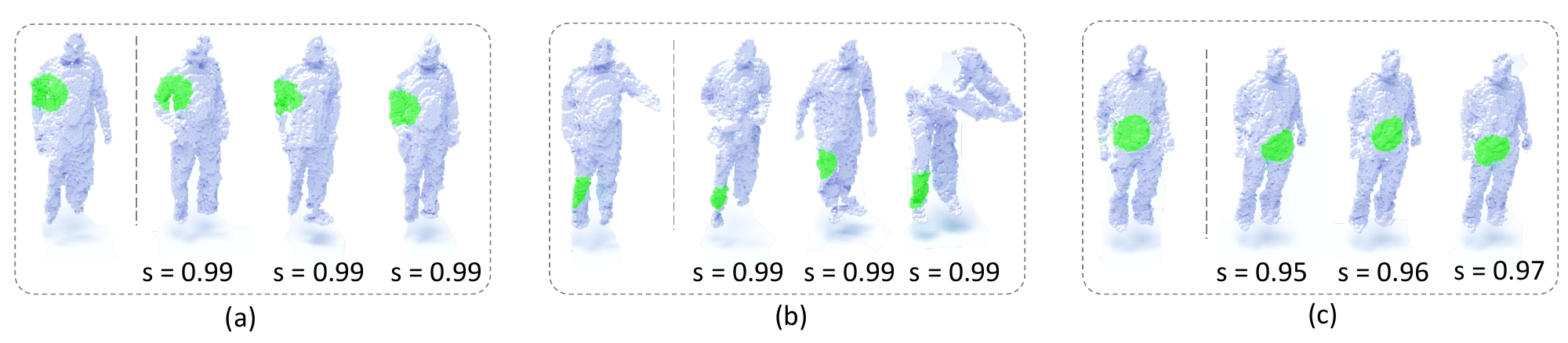}
        \caption{The visualization of positive neighbors. The neighbors with high similarities come from different actions or subjects. The number denotes the similarity.}
        \label{fig3}
\end{figure*}

\begin{table}[t]
    \centering
    \small 
    \caption{Gesture recognition accuracy (\%) on NvGesture.}
    \setlength{\tabcolsep}{4.5mm}
    \begin{tabular}{l|cc}
    \toprule
    \textbf{Methods} &\textbf{Input} &\textbf{NvGesture} \\
    \midrule
    FlickerNet~\cite{flickernet}           & point           & 86.3  \\
    PLSTM\small{-base}~\cite{min2020efficient}    & point    & 87.6 \\
    PLSTM\small{-early}~\cite{min2020efficient}   & point    & 93.5 \\
    PLSTM\small{-PSS}~\cite{min2020efficient}     & point    & 93.1 \\
    PLSTM\small{-middle}\cite{min2020efficient}   & point    & 94.7 \\
    PLSTM\small{-late}~\cite{min2020efficient}    & point    & 93.5 \\
    Kinet\cite{zhong2022no}                       & point    & 89.1         \\
    \rowcolor{gray!20}PSTNet\cite{pstnet} (50epochs)   & point            & 86.1       \\
    \midrule
    \textbf{PointCPSC} (50epochs)         & point            & 87.3       \\
    \bottomrule
    \end{tabular}
    \label{NvGesture}
\end{table}

\subsection{Ablation Studies}

We conduct ablation studies on MSRAction-3D and Synthia 4D.
On MSRAction-3D, 16-frame clips are sampled and the other hyper-parameters are the same as end-to-end fine-tuning.
On Synthia 4D, 4096-point frames are sampled and the models are finetuned for 75 epochs.
All hyper-parameters except for the ablated ones are kept the same for fair comparison.

\textbf{The Negatives with Appropriate Ratios.}
For current batch targets, we calculate their feature similarities with history embeddings stored in the memory bank.
We rank the similarities in descending order and use different ratios of embeddings as negatives.
The results are shown in Table~\ref{ratio}. 
It can be observed that our method achieves the highest accuracy with 70\% negatives and more negatives introduce moderate performance drops. 
{This indicates that there exist negatives with high similarity in the memory bank, namely undesired negatives, and they should be abandoned in pre-training.}

\textbf{The Utilization of Positive Neighbors.} 
Although the positive neighbors are retrieved based on feature similarity, how to utilize these neighbors still needs further exploration.
Three different schemes are compared and the results are presented in Table~\ref{scheme}.
Compared with integrating positive neighbors with feature similarity as softmax weight (B1), the accuracy of directly adding $K$ positive pairs (B2) has increased by 1\%.
When combining the target and its neighbors with their similarity, the performance is optimal.
By utilizing weight fusion, the comprehensive representations of positive samples are constructed and they are more generalized for performing contrastive learning.

\begin{table}[t]
    \centering
    \small
    \caption{The negatives with appropriate ratios.}
    \setlength{\tabcolsep}{1.9mm}
    \begin{tabular}{l|ccccc}
    \toprule 
      & \footnotesize{A1} & \footnotesize{A2 (Ours)} & \footnotesize{A3 } & \footnotesize{A4} \\
    \midrule
    \textbf{Negatives Ratio (\%)}     & 60 & 70 & 80  & 90   \\
    \textbf{Accuracy (\%)}  & 91.38 & \textbf{92.26} & 90.91  & 90.23  \\
    \bottomrule
    \end{tabular}
    \label{ratio}
\end{table}

\begin{table}[t]
    \centering
     \small
    \caption{Ablation study on positive neighbors.}
    \setlength{\tabcolsep}{2.9mm}
    \begin{tabular}{l|c|c}
    \toprule
       &   \textbf{Weighting Scheme}  & \textbf{Accuracy (\%)}\\
    \midrule
    \footnotesize{B1} &  Softmax Weighting      &   89.90      \\
    \footnotesize{B2} & Add $K$ positive pairs  &  90.91  \\ 
    \footnotesize{B3 (Ours)} & Feature Weighted Fusion                      &    \textbf{92.26}     \\
    \bottomrule
    \end{tabular}
    \label{scheme}
\end{table}

\begin{table}[t]
    \centering
    \small
    \caption{Results achieved using different numbers of positive neighbors.}
    \setlength{\tabcolsep}{2.9mm}
    \begin{tabular}{l|ccccc}
    \toprule 
      & \footnotesize{C1} & \footnotesize{C2 (Ours)} & \footnotesize{C3 } & \footnotesize{C4} \\
    \midrule
    \textbf{Numbers}     & 1 & 3 & 5  & 10  \\
    \textbf{Accuracy (\%)}  & 91.58 & \textbf{92.26} & 91.25  & 90.57  \\
    \bottomrule
    \end{tabular}
    \label{neighbors number}
\end{table}

\textbf{The Number of Positive Neighbors.}
The highly similar neighbors are mined from other instances as positive supplements.
We evaluate the number of positive neighbors and the results are shown in Table~\ref{neighbors number}.
As we can see, the performance is improved as the number of positive neighbors is increased from 1 to 3. However, further increase of positive neighbors cannot introduce accuracy gains but lead to lower performance. Consequently, 3 positive neighbors are used as the default setting in our experiments.

\textbf{The Size and Cost of Memory Bank.}
We investigate the performance of memory banks with different sizes in terms of running time and memory consumption. Specifically, we evaluate models with the memory bank size of 256, 512, and 1024.
The results are shown in Table~\ref{size}.
It can be observed that our method achieves higher accuracy with a larger memory bank, producing an accuracy of 92.26\% with a  memory bank of size 1024.
Meanwhile, running time and memory consumption have an acceptable increase.
Consequently, 1024 is used as the default size of the memory bank in our experiments.

\begin{figure*}[ht]
        \centering
        \includegraphics[width=1\linewidth]{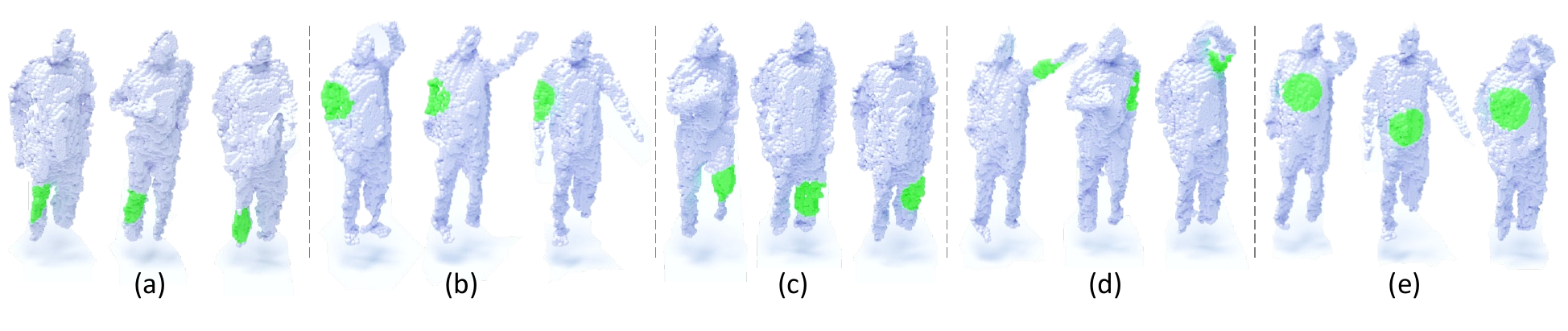}
        \caption{The visualization of the prototypes learned in pre-training. Different prototypes correspond to specific human body regions. }
        \label{fig4}
\end{figure*}

\begin{table}[t]
    \centering
    \small
    \caption{Time (mins/epoch), memory (MiB), and accuracy (\%) achieved using memory banks with different sizes.
    }
    \setlength{\tabcolsep}{2.6mm}
    \begin{tabular}{l|c|c|c|c}
    \toprule
    & \textbf{Size}  & \textbf{Time}  & \textbf{Memory} & \textbf{Accuracy (\%)}\\
    \midrule
    \footnotesize{D1} &  256  &   1.2          &   8647               &   90.91      \\
    \footnotesize{D2} &  512  &   1.7          &    9435              &   91.57      \\
    \footnotesize{D3 (Ours)} &  1024 &  2.1     &  10276  & \textbf{92.26}  \\
    \bottomrule
    \end{tabular}
    \label{size}
\end{table}

\begin{table}[t]
    \centering
    \small
    \caption{Accuracy (\%) achieved using different numbers of prototypes.}
    \setlength{\tabcolsep}{2.6mm}
    \begin{tabular}{l|c|c|c}
    \toprule
    & \textbf{Prototypes}  & \textbf{MSRAction-3D}  & \textbf{Synthia 4D} \\
    \midrule
    \footnotesize{E1 (Ours)} &  10  &  \textbf{92.26}       &  \textbf{71.13}      \\
    \footnotesize{E2} &  20  &     90.91                   &   70.53      \\
    \footnotesize{E3} &  30  &       90.91                &   70.28      \\
    \bottomrule
    \end{tabular}
    \label{clusters}
\end{table}

\textbf{The Size of Prototypes on Different Benchmarks.}
We also study the number of prototypes on different benchmarks.
The results are shown in Table~\ref{clusters}.
Our method achieves the highest accuracy on MSRAction-3D with the prototype number of 10.
Intuitively, the prototypes aggregated from superpoint representations can be viewed as human body parts with specific semantics.
From this point of view, introducing too many prototypes may suffer semantic-less fragments and decrease the performance.
On Synthia 4D, our method achieves the highest accuracy of  71.13\%. 
It maybe because this prototype number is close to the object categories in this dataset. This indicates that a suitable number of prototypes that fit the dataset well is beneficial to the final performance.

\begin{figure}[t]
	\centering
	\includegraphics[width=1\columnwidth]{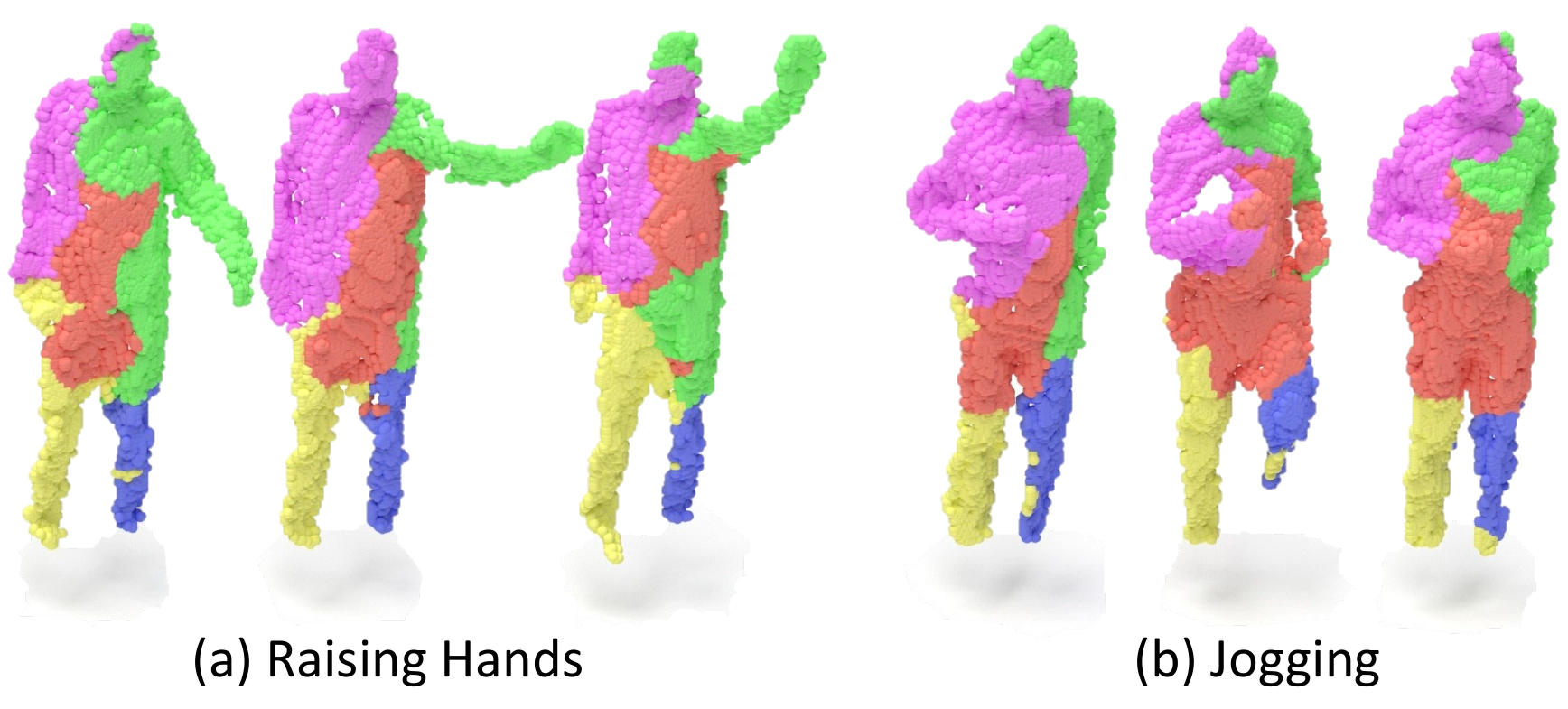}
	\caption{The visualization of human motion segmentation.
 }
	\label{fig5}
\end{figure}


\textbf{The Effectiveness of Self-supervised Tasks.} 
We evaluate the effectiveness of local contrastive prediction, positive neighbors, and the pretext task of semantic clustering on different datasets.
The results are shown in Table~\ref{tasks}.
Note that, we only perform prototype alignment on MSRAction-3D.
On MSRAction-3D, the supplements of positive neighbors with the local contrastive prediction branch improve the accuracy to 91.98\%. 
When prototype alignment is introduced, our method achieves an accuracy of 92.26\%.
On Synthia 4D, the retrieved positive neighbors also contribute to performance improvement.
{More importantly, joint prototype alignment and soft category alignment further improve the segmentation accuracy with a large margin.}
This demonstrates that our method can well capture fine-grained cues that benefit the semantic segmentation.


\subsection{Qualitative Analysis}

\textbf{Positive Neighbors.} During pre-training, several highly similar superpoints stored in the memory bank are selected based on feature similarity.
The corresponding raw points areas of superpoints are visualized in Fig.~\ref{fig3}.
These superpoints come from diverse videos and categories, but present highly similar human body regions.
Since our self-supervised pre-training aims to model local dynamics, these highly similar superpoints from other instances should be treated as positive neighbors.
This motivates us to design the strategy of sample selection, to achieve effective contrast and learn robust representations.

\textbf{Prototypes Visualization.} We explore what the prototypes have learned by classifying the superpoints aggregated from raw point cloud sequences with pre-trained prototypes.
 We randomly select several prototypes and evaluate four videos.
The visualization results are shown in Fig.~\ref{fig4}.
Each prototype corresponds to a specific region of human bodies.
This demonstrates that our self-supervised framework effectively models local structures and learns high-level semantics beneficial for downstream tasks.

\textbf{Potential Applications.}
We visualize the learned prototypes on two point cloud sequences in Fig.~\ref{fig5}, where each color represents a prototype.
When visualizing, the prototypes with the same semantics are incorporated.
It can be seen that these pre-trained prototypes embed specific human body parts.
This demonstrates that the pretext task of semantic clustering models human parts from superpoint representations.
Intuitively, the prior information learned in pre-training is beneficial for non-rigid motion segmentation.
Besides, the soft category assignments of points may benefit interactive annotation tasks.

\begin{table}[t]
    \centering
    \small
    \caption{Ablation results on different benchmarks.}
    \setlength{\tabcolsep}{1.9mm}
    \begin{tabular}{l|c|c|c}
    \toprule
    & \textbf{Tasks}  & \textbf{MSR (\%)} & \textbf{Syn (\%)}   \\
    \midrule
    \footnotesize{F1}    &    Local Contrastive Prediction       &    91.38         &  70.01     \\
    \midrule
    \footnotesize{F2}    &    F1 + Sample Selection Strategy     &    91.98         &    70.45     \\
    \midrule
    \footnotesize{F3}   &  F2 + Prototype Alignment  &   \textbf{92.26}  &  70.73   \\
    \midrule
    \footnotesize{F4}   &  F2 + Soft Category Alignment  &   -  &  70.67 \\
    \midrule
    \footnotesize{F5}   &  \makecell[c]{ F2 + Prototype Alignment \\ + Soft Category Alignment}  &  -   &  \textbf{71.13}   \\
    \bottomrule
    \end{tabular}
    \label{tasks}
\end{table}

\section{Conclusions}
We propose a unified self-supervised framework for pre-training on point cloud videos.
To adapt the self-supervised framework for diverse object-centric and scene-centric downstream tasks, we design the pretext task of semantic clustering, which achieves hierarchical semantic alignment between predictions and targets.
In addition, we retain proper negatives for effective contrast and select highly similar negatives as positive neighbors for robust representations. 
Extensive experiments and ablation studies are performed to demonstrate the effectiveness of our self-supervised framework.

\noindent\textbf{Acknowledgments.} This work was partially supported by the Fundamental Research Funds for the Central Universities (No.226-2023-00048), the National Natural Science Foundation of China (No.61673270, 61973212, 61972435, 61602499), Artificial Intelligence Key Laboratory of Sichuan Province (2022RZY02), and Guangdong Basic and Applied Basic Research Foundation (2019A1515011271).

{\small
\bibliographystyle{ieee_fullname}
\bibliography{egbib}
}

\end{document}